\documentclass[10pt,twocolumn,letterpaper]{article}

\usepackage[pagenumbers]{wacv} % To force page numbers, e.g. for an arXiv version

\newcommand{\red}[1]{{\color{red}#1}}

\definecolor{wacvblue}{rgb}{0.21,0.49,0.74}
\usepackage[pagebackref,breaklinks,colorlinks,allcolors=wacvblue]{hyperref}

\def\wacvPaperID{371} % *** Enter the WACV Paper ID here
\def\confName{WACV}
\def\confYear{2026}

\PassOptionsToPackage{table,dvipsnames}{xcolor}
\usepackage{xcolor}
\usepackage{colortbl}   % defines \rowcolor and \cellcolor

\usepackage{booktabs}   % defines \toprule, \midrule, \bottomrule

\usepackage{graphicx}   % defines \resizebox
\usepackage{booktabs}
\usepackage{graphicx}      % for \resizebox
\usepackage{booktabs}
\usepackage{graphicx}
\usepackage{array}
\usepackage{bbding}
\usepackage{pifont}
\usepackage{wasysym}
\usepackage{amssymb}
\usepackage{multirow}
\usepackage{multicol}
\usepackage{makecell}
\usepackage{caption}
\usepackage{wrapfig}
\usepackage{enumitem}
\usepackage{amsmath} 
\usepackage[capitalize]{cleveref}
\usepackage{hyperref}
\usepackage{changepage}
\usepackage{float}
\usepackage{float}            % for [H] float placement specifier
\usepackage{graphicx}         % for \resizebox
\usepackage{tcolorbox}        % for \tcbox
\usepackage{tabularray}       % for \tblr

\newcommand{\myparagraph}[1]{\vspace{5pt}\noindent{\bf #1}}
\title{Scene Graph-Guided Proactive Replanning for Failure-Resilient Embodied Agents}
\vspace{-1.5em}
\author{
  \hspace{-0.3cm}
  Che Rin Yu\textsuperscript{1} \quad
  Daewon Chae\textsuperscript{1} \quad
  Dabin Seo\textsuperscript{1} \quad
  Sangwon Lee\textsuperscript{2} \quad
  Hyeongwoo Im\textsuperscript{2} \quad
  Jinkyu Kim\textsuperscript{1}\\
  \textsuperscript{1}Korea University \quad
  \textsuperscript{2}KT (Korea Telecom) R\&D Center \\
  \texttt{\{eyucherin, cdw098, lemonstar99, jinkyukim\}@korea.ac.kr} \\
  \texttt{\{lee.sangwon, im.hyeongwoo\}@kt.com}
}

\begin{document}
\maketitle
\begin{abstract}
When humans perform everyday tasks, we naturally adjust our actions based on the current state of the environment. For instance, if we intend to put something into a drawer but notice it is closed, we open it first. However, many autonomous robots lack this adaptive awareness. They often follow pre-planned actions that may overlook subtle yet critical changes in the scene, which can result in actions being executed under outdated assumptions and eventual failure. While replanning is critical for robust autonomy, most existing methods respond only after failures occur, when recovery may be inefficient or infeasible. Although proactive replanning holds promise for preventing failures in advance, current solutions often rely on manually designed rules and extensive supervision. In this work, we present a proactive replanning framework that detects and corrects failures at subtask boundaries by comparing scene graphs constructed from current RGB-D observations against reference graphs extracted from successful demonstrations. When the current scene fails to align with reference trajectories, a lightweight reasoning module is activated to diagnose the mismatch and adjust the plan. Experiments in the AI2-THOR simulator demonstrate that our approach detects semantic and spatial mismatches before execution failures occur, significantly improving task success and robustness. Our code will be publicly available upon publication.
\end{abstract}
\section{Introduction}
\label{sec:intro}

\begin{figure}[t]
    \centering
    \includegraphics[width=1.0\linewidth]{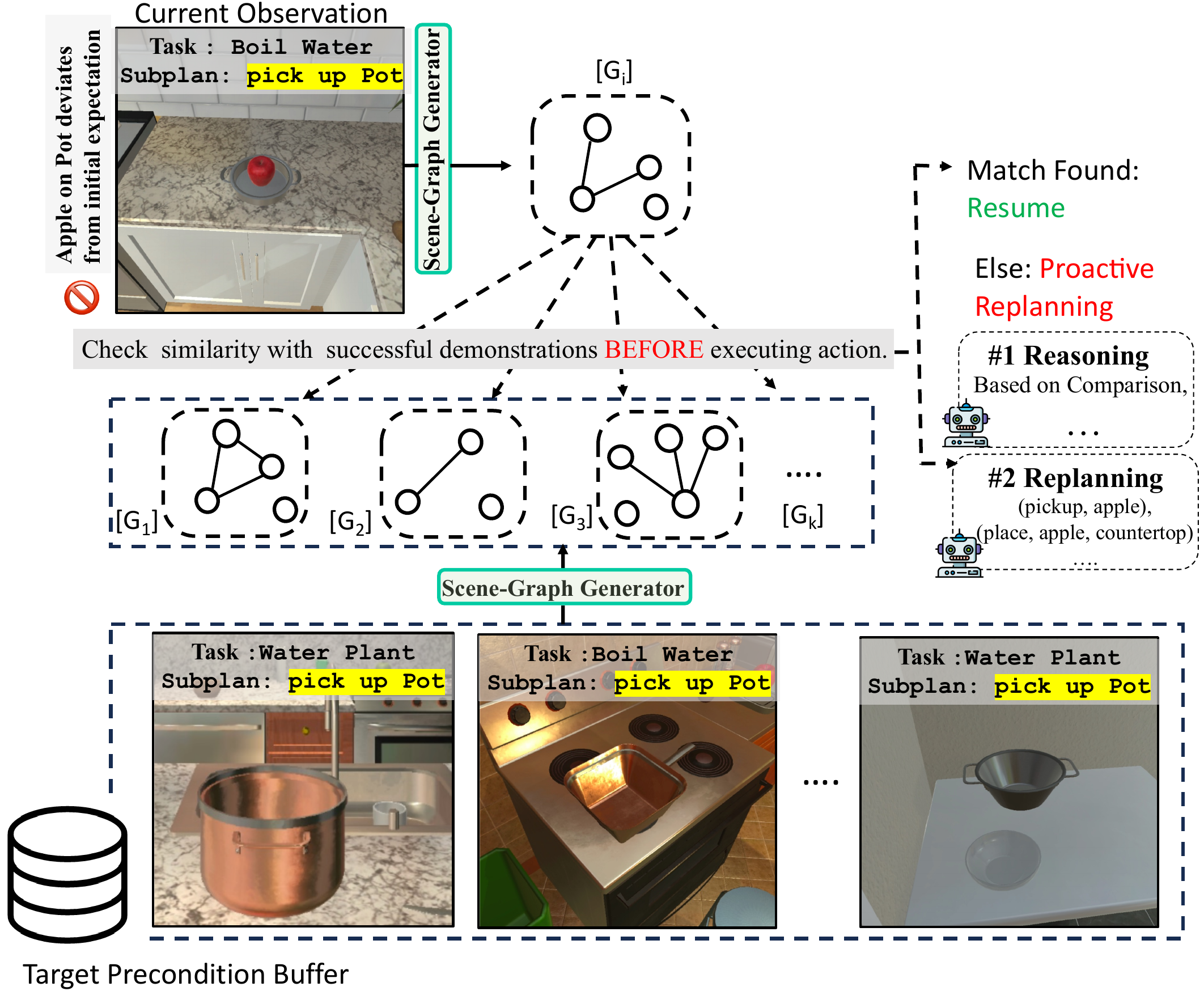}
    \caption{\textbf{Overview our framework.}  Before executing each subtask, the system generates a scene graph of the current observation and compares it to previous successful demonstrations. If a sufficiently similar match is found, execution proceeds as planned. Otherwise, the agent proactively triggers replanning by reasoning over the scene discrepancy (e.g., detecting that an apple on the pot deviates from expected conditions).}
\vspace{-1.5em}
    \label{fig:small-teaser}
\end{figure}

\begin{figure*}[t]
    \centering
    \includegraphics[width=1.0\linewidth]{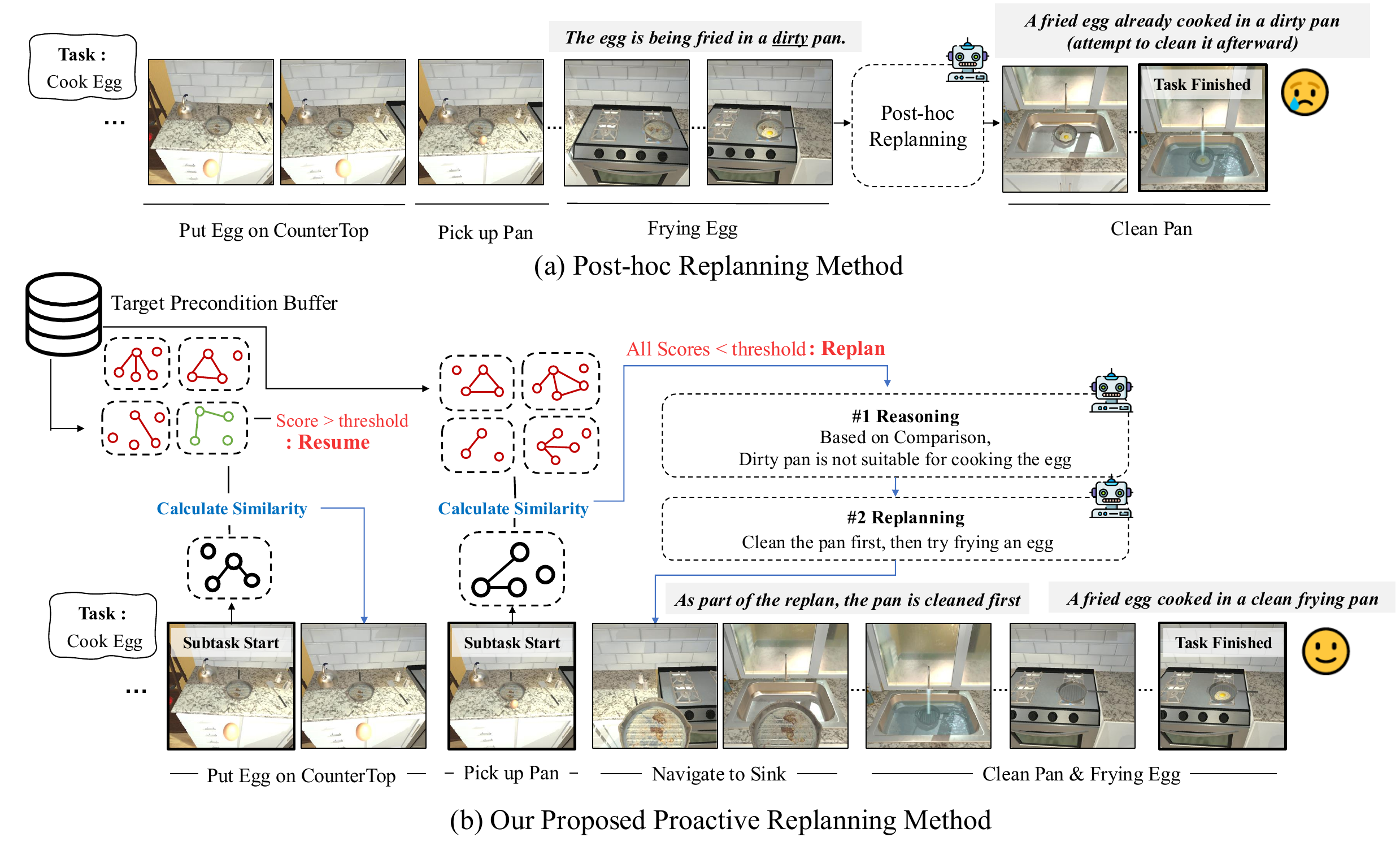}
    % \vspace{-2.0em}
    \caption{(a) An example of a conventional post-hoc replanning strategy where an agent reacts only after an undesired outcome has occurred (or an irreversible failure has taken place). For instance, the agent detects that an egg is being fried in a dirty pan and then takes corrective action by replanning. In contrast, (b) our proposed proactive replanning strategy enables an agent to continuously monitor precondition states and proactively replans subsequent actions to prevent potential failures. For example, the agent recognizes in advance that a dirty pan is unsuitable for cooking and adjusts its plan to clean the pan before proceeding. 
    %(1) illustrates a post-hoc method, where the robot detects that the pan is dirty only after completing the task, resulting in an irreversible failure. In contrast, (2) illustrates our approach: by comparing the observed environment to an target precondition state before picking up the pan, the system detects the discrepancy early and replans to clean it first, preventing failure before it materializes.
    }
    \label{fig:teaser}
    \vspace{-1.5em}
\end{figure*}
% Autonomous robots are increasingly capable of performing complex tasks, promising transformative applications in unstructured real-world environments~\citep{fu2024mobile, opensource_soft_robotic,learning_to_manipulate}. However, to operate reliably, robots must dynamically adapt their behavior in response to unexpected environmental changes, errors\jinkyu{what errors?}, or failures that arise during task execution. This makes replanning, \jinkyu{definition of replanning}, a fundamental requirement for robust autonomy, particularly in unpredictable settings where failures may involve irreversible or safety-critical consequences that cannot be mitigated through simple recovery mechanisms. 

Autonomous robots are increasingly capable of performing complex tasks, promising transformative applications in unstructured real-world environments~\citep{fu2024mobile, opensource_soft_robotic,learning_to_manipulate}. However, most existing robotic systems carry out actions according to strategies formulated at planning time, frequently overlooking environmental changes and acting on outdated assumptions, leading to unsafe or irreversible failures. This makes replanning, the process of revising or generating a new plan based on updated observations, a fundamental requirement for robust autonomy, particularly in unpredictable environments.   

However, building robust replanning capabilities for autonomous systems remains fundamentally challenging, requiring the three following components:  (\textit{i}) determining the optimal time for replanning interventions in an efficient manner, (\textit{ii}) accurately diagnosing the root causes of failures or potential failure conditions, and (\textit{iii}) generating effective and corrective action sequences that can recover progress toward the task goal. Successfully addressing these components is essential for enabling robots to operate reliably across diverse environments.

% These challenges are further compounded in long-horizon tasks, where failures often arise not only from semantic misunderstandings, but also from subtle, often overlooked assumptions about spatial configurations~\citep{kamath2023s,shiri2024empirical,yamada2023evaluating}. For instance, whether an object is properly placed, partially obstructed, or held can critically affect subtask feasibility, where such spatial arrangements may not be immediately apparent from semantic information alone. As a result, robust failure reasoning and analysis require not only high-level symbolic abstraction, but also fine-grained spatial grounding that accurately captures both the relational and geometric structure of the environment. This motivates the development of structured spatial representations capable of supporting reliable detection, reasoning, and replanning in response to failures.

These challenges are further compounded in long-horizon tasks, where failures often stem not only from incorrect semantic misunderstandings but also from overlooked spatial configurations about the scene~\citep{kamath2023s,shiri2024empirical,yamada2023evaluating}. For example, whether an object is partially visible, obstructed, or held by the agent can directly affect task feasibility. This motivates the development of structured scene representations that capture both relational and geometric context, enabling more accurate failure detection and informed replanning.

Although recent studies have made significant progress in tackling the challenges of robotic failure recovery, many approaches rely mainly on post-hoc mechanisms (i.e., responding only after failures emerge)~\citep{reflect,guo2024doremi, sarch2023openended,wang2024llm}, predefined rule-based triggers~\citep{yang2024text2reaction, raman2024cape, kim2024preemptive}, or expensive human supervision~\citep{zha2023distilling, DBLP:journals/corr/abs-2403-12910, liu2023interactive, cui2023no}. Consequently, they struggle to proactively and autonomously prevent failures. Rule-based methods, in particular, often rely on rigid, hardcoded conditions that overlook the richness of visual context. A key foundation for such capabilities is the concept of preconditions in robotic task planning, which refer to the predefined conditions that must be satisfied before an action can be successfully executed (e.g., a pan must be clean before use, or a container must be empty before placing an object inside). While preconditions are often defined manually through hand-engineered rules, this method tends to break down in the face of visually diverse conditions. To address this limitation, we take inspiration from how humans assess their environment. People intuitively evaluate their surroundings by drawing on accumulated perceptual experience. For instance, before opening a drawer, we don’t consciously verify a checklist of conditions, we simply look at it and recognize whether it’s already open, blocked, or otherwise inaccessible. This judgment relies on internalized visual cues, allowing us to quickly determine whether an action is feasible 

% This judgment isn’t made through explicit logical reasoning but rather through internalized visual patterns developed over time. These perceptual cues enable us to quickly and reliably assess whether it’s appropriate to proceed with an action under the current conditions.

% Rather than relying on fixed rules, humans use learned experiences to interpret visual scenes and assess whether the conditions are suitable for taking action. For instance, when attempting to open a drawer, we don’t rely on step-by-step logic, we instinctively determine if it’s feasible by recognizing familiar visual patterns like it being open or obstructed through prior experiences. These learned perceptual cues enable us to anticipate whether an action is likely to succeed in the current context. 

\begin{figure}[t]
    \centering
    \vspace{-0.5em} % fine-tune as needed
    \includegraphics[width=0.9\linewidth]{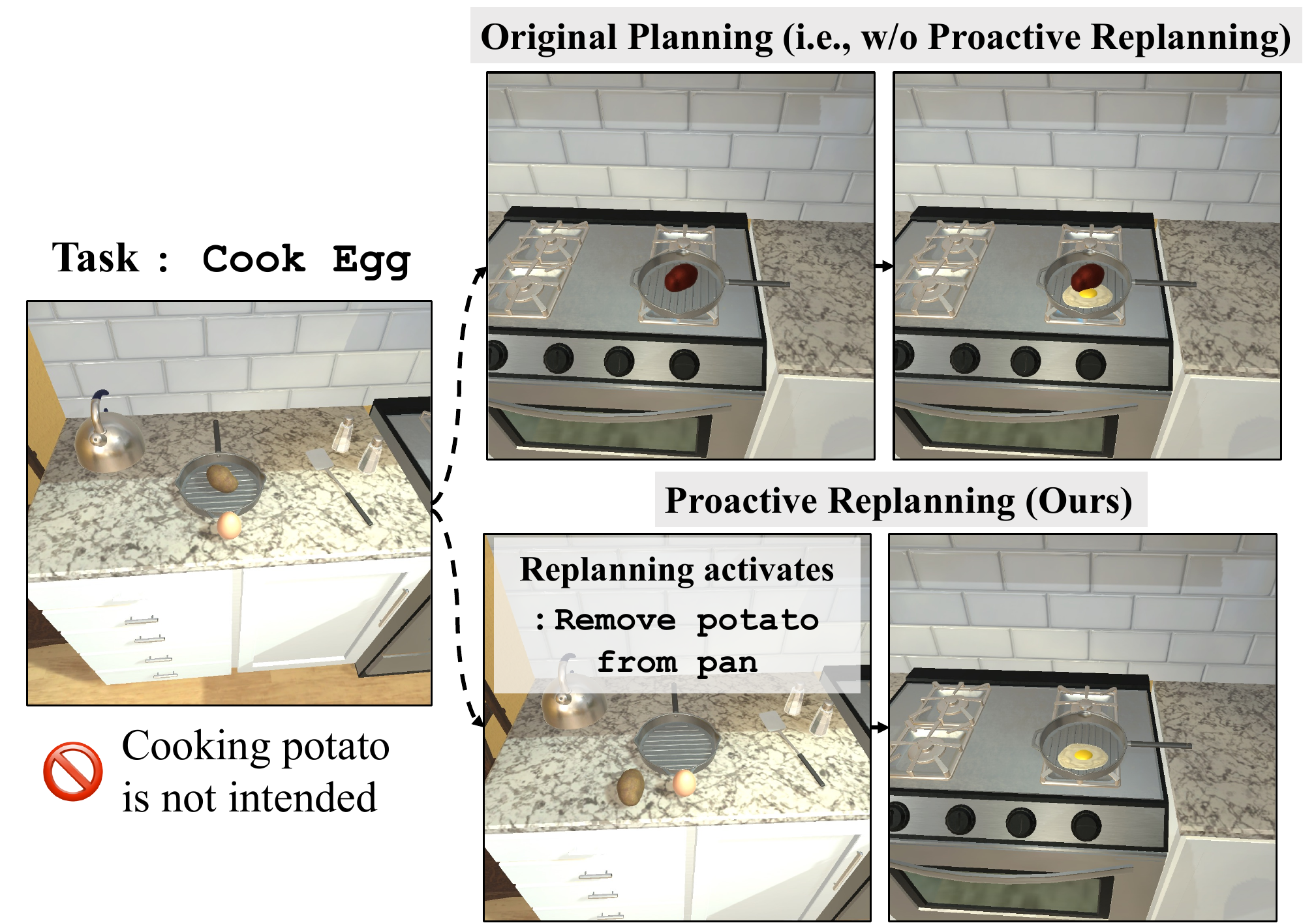}
    \caption{To cook an egg, our method (bottom) replans early to remove the potato. The original plan (top) does not consider the potato, resulting in both being cooked.}
    \label{fig:proactive_ex}
    \vspace{-2em}
\end{figure}

Building on this idea, our method proactively detects potential failures before action execution and dynamically revises plans by grounding decisions in visual scene understanding and successful demonstrations of long-horizon tasks (e.g., boiling water, cooking an egg), as shown in Figures~\ref{fig:small-teaser},~\ref{fig:teaser}, and ~\ref{fig:proactive_ex}. This approach contrasts with conventional post-hoc replanning strategies that address failures only after they occur (compare Figure~\ref{fig:teaser} (a) and (b)). Specifically, at the beginning of each sub-task (e.g., ``picking up a pan''), the agent compares the scene graph of the current environment with expected scene graphs derived from prior successful demonstrations. If the similarity falls below a predefined threshold, the system proactively triggers replanning before executing the subtask. This process generates a reasoning chain to identify the likely cause of failure (e.g., ``a dirty pan is unsuitable for cooking'') and subsequently formulates corrective actions (e.g., ``clean the pan before frying the egg''), thereby avoiding the predicted failure and enabling successful task completion. We validate the effectiveness of our method using the AI2-THOR simulator~\citep{kolve2017ai2}, demonstrating the agent's ability to preemptively avoid or recover from potential failures during task execution. Our contributions are as follows: 
\begin{itemize}[leftmargin=1.5em]
    \item We propose a novel proactive replanning approach that preemptively identifies and mitigates potential failures to ensure reliable achievement of target objectives in long-horizon tasks.
    %We introduce a novel proactive replanning method that preemptively identifies and corrects failures by leveraging target precondition states as predictive references for real-time failure analysis. 
    \item We present a lightweight scene graph-based failure anticipation method that leverages a structural visual understanding of the environment to assess, at the onset of each sub-task, whether the current scene satisfies the necessary conditions for successful action execution.
    %We present a lightweight, scene graph-based potential failure detection algorithm that efficiently identifies precondition mismatches and triggers corrections through graph comparisons. 
    \item We demonstrate the effectiveness of our approach, showing that it not only improves failure detection and task success rates compared to baseline methods, but also enhances failure reasoning quality by grounding decisions in visual and spatial cues, as validated through human evaluations.
\end{itemize}

\section{Related Work}
\label{sec:related_work}

\begin{figure*}[t]
    \centering
    \includegraphics[width=1.0\linewidth]{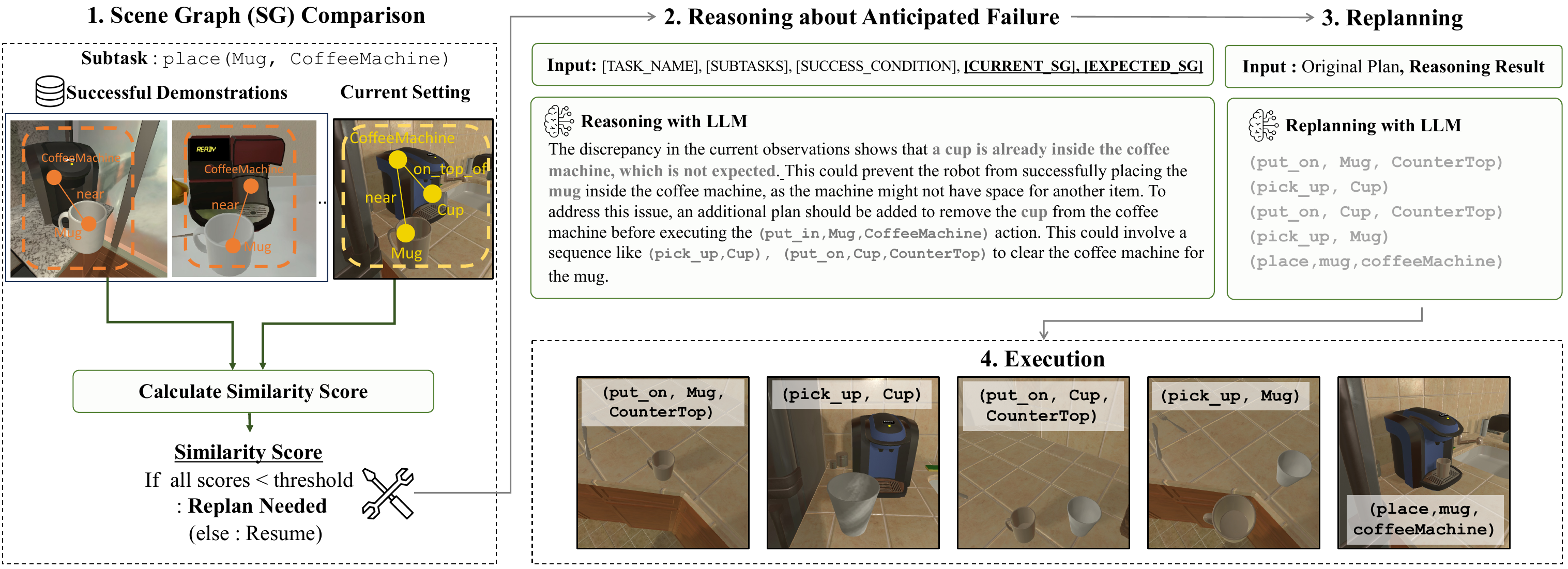}
    % \vspace{-2.0em}
    \caption{
    Our proposed method comprises four main steps: (1) We compute scene graph-based similarity scores between the current and expected scenes, using a database of successful demonstrations, to assess subtask feasibility. (2) If the score falls below a predefined threshold, a large language model (LLM) generates potential failure reasons based on the current scene. (3) Using the original plan and the inferred reasoning, we generate corrective actions to avoid or mitigate the failure. (4) The agent then executes the revised plan.
    }
    \label{fig:overall}
     \vspace{-1.5em}
\end{figure*}

% \subsection{Strategies for Robotic Replanning}
\myparagraph{Strategies for Robotic Replanning}. There has been a growing interest in replanning strategies in the robotics community where approaches have been introduced based on predefined rule-based triggers~\citep{yang2024text2reaction, raman2024cape, kim2024preemptive, cornelio2024recover}, human-in-the-loop strategies~\citep{zha2023distilling, DBLP:journals/corr/abs-2403-12910, cui2023no}, post-hoc replanning strategies~\cite{reflect,guo2024doremi, sarch2023openended,wang2024llm}, and large vision-language model-based methods~\cite{duan2025aha, dai2024racerrichlanguageguidedfailure}. Human-in-the-loop~\citep{zha2023distilling, DBLP:journals/corr/abs-2403-12910, cui2023no} strategies offer operational flexibility by allowing supervisory intervention during task execution, but they introduce scalability challenges, which are impractical for fully autonomous operation, and impose significant labor costs. Post-hoc replanning methods~\citep{reflect,guo2024doremi, sarch2023openended,wang2024llm}, which analyze completed subtasks or trajectories to identify failure points, are shown effective but unsuitable when corrective action must occur before a failure materializes. Moreover, these approaches inherently struggle with irreversible failures where retrospective corrections are infeasible. Lastly, more recent methods leverage vision-language models trained on failed trajectories~\citep{duan2025aha, dai2024racerrichlanguageguidedfailure} show promise in recognizing failure scenarios, but require  large-scale annotations, incure data collection costs, and often fail to generalize to rare or unseen failure modes.

\myparagraph{Large Language Models (LLMs) in Robotics.} 
Multi-modal large language models (LLMs) have been widely used in robotic task planning~\citep{ahn2022can,huang2022inner,huang2022language,rana2023sayplan,huang2022language}, translating high-level instructions into action sequences or executable code ~\citep{singh2023progprompt,liang2023code}. Similarly, Vision-Language Models (VLMs) interpret visual scenes~\citep{pan2004automatic,farhadi2010every,ordonez2011im2text,yao2010i2t} and detect task-relevant objects~\citep{redmon2016you,girshick2014rich} to extract visual information and align semantic cues in the environment. However, VLMs often overlook geometric and precise spatial structures, and struggle with occlusions, spatial constraints, and geometric plausibility in complex scenes~\citep{kamath2023s,shiri2024empirical,yamada2023evaluating}. Moreover, LLMs and VLMs may hallucinate~\citep{zhou2023analyzing, zhai2023halle} and cause errors while assessing environment states, where minor spatial inconsistencies such as occluded objects, misplaced items, or obstructed targets, may silently interfere with task execution. Solely relying on semantics may not detect such failures, highlighting the need for explicit spatial reasoning to detect discrepancies in real time. To address this, we propose a novel framework combining symbolic scene graph comparisons with selective LLM reasoning for proactive failure detection, while also efficiently leveraging LLMs with minimal inference overhead.

\section{Method}
\label{sec:method}

\label{sec:method_failure_detection}

\myparagraph{Problem Formulation.} We consider the problem of executing long-horizon tasks, each defined by a desired success condition $C_{\text{goal}}$ that describes the final state the robot must achieve (e.g., ``a mug filled with coffee is placed on the table''). To achieve $C_{\text{goal}}$, a task $\mathcal{T}$ is decomposed into an ordered sequence of $n$ high-level subtasks, $\mathcal{T} = [a_1, a_2, \dots, a_n]$, where each subtask $a_i$ corresponds to a semantically meaningful action (e.g., ``grab mug,''). Each subtask is intended to transition the environment closer to satisfying $C_{\text{goal}}$, such that the full execution of $\mathcal{T}$ leads to potential  task completion under expected conditions. 

% At execution time, the robot receives an RGB-D observation $I_i$ of the environment immediately before executing each subtask $a_i$, from which it extracts a structured semantic representation in the form of a scene graph $G_i$ that captures the entities and their relations present at that moment. To assess task progress and detect potential failures, the robot compares $G_i$ to the expected scene graph $\hat{G}_i$ inferred from the target precondition state $P_i$, which may be derived from a successful (i.e., reference) trajectory, manual labels from human annotation, or descriptions generated by large language models (LLMs) based on commonsense knowledge. In our setting, we leverage a dataset of reference trajectories $\mathcal{D}$, where each trajectory $\tau \in \mathcal{D}$ specifies how a high-level task $\mathcal{T}$ should proceed under expected conditions, including the final success condition $C_{\text{goal}}$ and the ordered sequence of subtasks $[a_1, a_2, \dots, a_n]$ required for completion. From these trajectories, we extract the expected precondition state for each subtask, which serves as an intrinsic reference for comparison during execution.

At execution time, the robot receives an RGB-D observation $I_i$ of the environment immediately before executing each subtask $a_i$. From this multimodal input, it constructs a structured semantic representation in the form of a scene graph $G_i$, which captures not only the entities present but also their spatial and relational configurations as perceived through visual cues. To evaluate the feasibility of executing $a_i$, the robot compares $G_i$ to expected scene graphs $\{\hat{G}_i^1, \hat{G}_i^2, \dots, \hat{G}_i^k\}$, each representing valid configurations observed at the same subtask step across $k$ successful demonstrations, without necessarily originating from the same overall task that the robot is currently executing. If the current scene graph $G_i$ does not closely match the expected graphs beyond a certain similarity threshold, the system considers the current environment misaligned with any known successful condition and proactively triggers replanning to prevent likely failure. This strategy enables the robot to adapt to a variety of valid but visually distinct task configurations observed across a set of demonstrations.

\begin{figure}[t]
    \centering
    \vspace{-0.5em} % fine-tune as needed
    \includegraphics[width=1.0\linewidth]{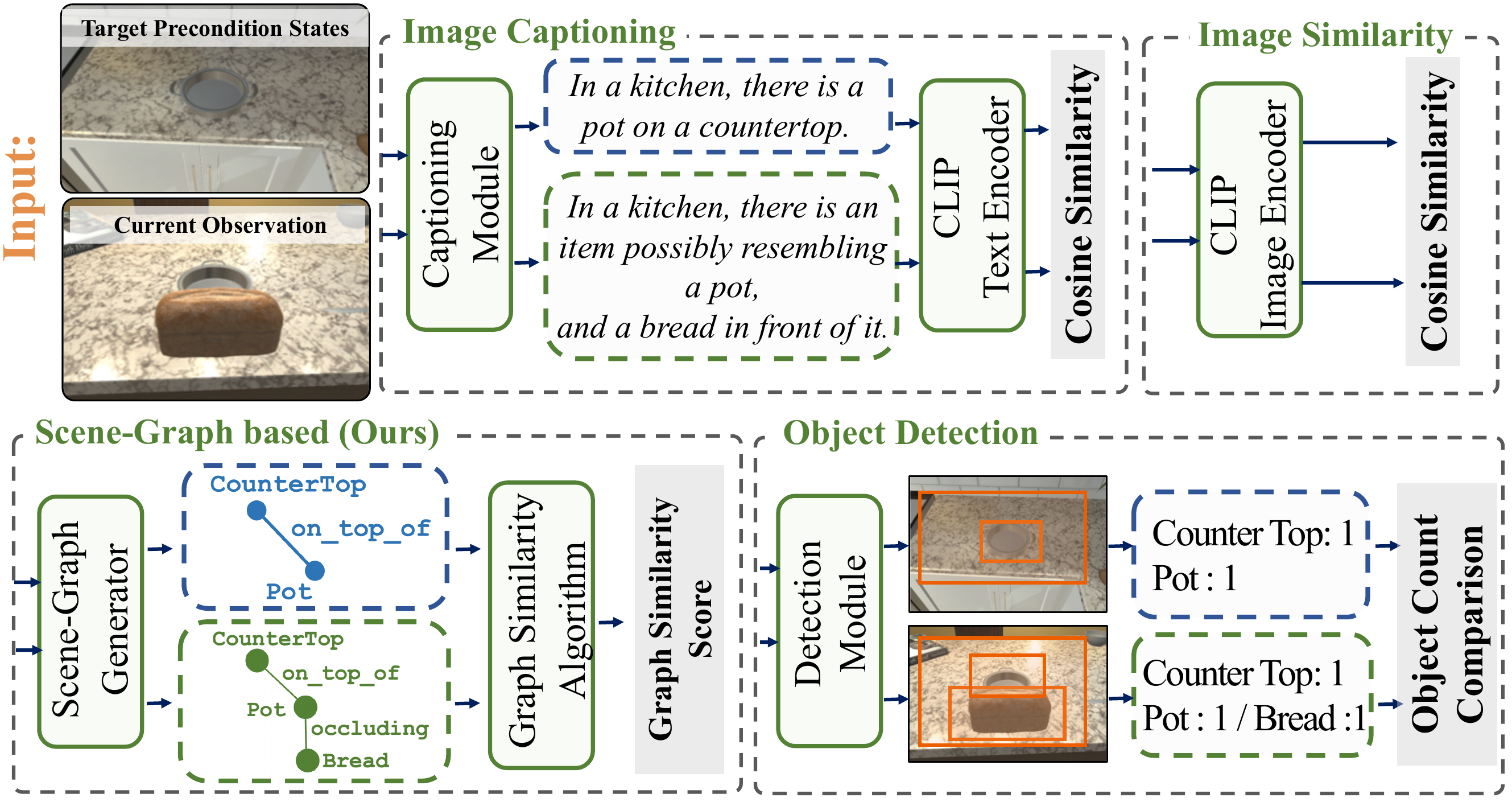}
    \caption{llustration of four potential approaches for detecting failures prior to subtask execution:
(1) \textbf{Image Similarity}-based model computes CLIP-based similarity between current and expected RGB images. (2) \textbf{Image Captioning}-based model compares text descriptions generated from each image using CLIP text embeddings. (3) \textbf{Object Detection}-based model compares object counts from current and expected scenes. (4) Our \textbf{Scene Graph}-based model performs graph-based discrepancy analysis using structured scene representations.}
    \label{fig:potential_detection}
     \vspace{-1.5em}
\end{figure}

\myparagraph{Graph-Based Failure Detection}. As illustrated in Figure~\ref{fig:potential_detection}, we explore several alternative approaches for detecting failures prior to subtask execution: (i) \textit{image-level comparison}, which uses CLIP~\citep{clip} embedding similarity between current and reference RGB images; (ii) \textit{caption-based comparison}, which generates scene descriptions using vision-language models and compares them in the embedding space; and (iii) \textit{object-level matching}, which compares object category counts from detected bounding boxes.

Each method offers certain advantages, image similarity is lightweight, captioning captures higher-level semantics, and object detection provides symbolic grounding. However, they also present key limitations: image similarity is overly sensitive to visual noise, captioning introduces linguistic ambiguity, and object matching lacks spatial context. These drawbacks hinder their reliability in identifying failures rooted in relational or structural mismatches. In contrast, our scene graph-based approach captures not only object presence but also spatial configurations and semantic relations critical to subtask feasibility. Further details on the baseline methods and their limitations can be found in the appendix.

\label{sec:method_failure_detection}
\myparagraph{3D Scene Graph Construction}. To represent the robot's semantic understanding of the environment prior to executing each subtask, we construct a visually grounded, task-informed scene graph from the robot’s RGB-D observation. Each scene graph captures objects, their states, and spatial relationships that are relevant to the current subtask context. Our approach is inspired by the scene graph construction methods proposed in REFLECT~\citep{reflect}, which summarize multimodal sensory inputs into symbolic structures for post-hoc failure explanation, and RoboEXP~\citep{roboexp}, which introduces an action-conditioned 3D scene graph that models interactive and spatial relations between objects and actions.

\begin{figure}[t]
    \centering
    \vspace{-0.5em} % fine-tune as needed
    \includegraphics[width=1.0\linewidth]{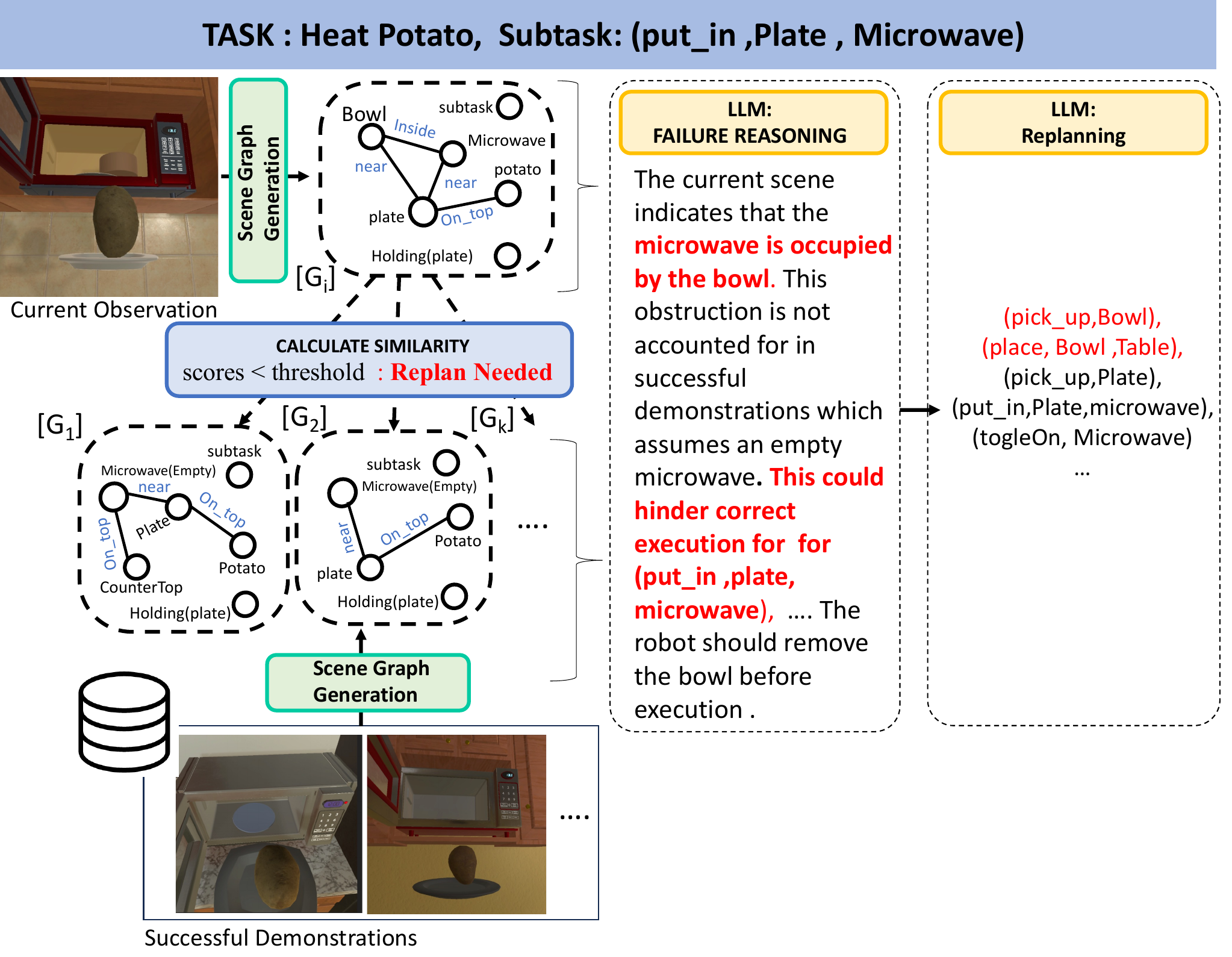}
    \caption{Example of proactive replanning triggered before executing the subtask \texttt{(put\_in, Plate, Microwave)} in the \textit{Heat Potato} task. The system detects that the microwave is already occupied by a bowl. By comparing scene graphs and reasoning over discrepancies, the agent identifies this obstruction as a potential failure point and proactively replans to remove the bowl before proceeding.}
    \label{fig:scene_graph_ex}
    \vspace{-1.5em}
\end{figure}

Rather than generating scene graphs at every frame, we construct one at the beginning of each subtask to evaluate whether the current environment is appropriate for continuing with the action, as shown in Figure~\ref{fig:scene_graph_ex}. Given the RGB-D image, we apply an object detector to identify bounding boxes and semantic segmentations, and crop object regions to infer state attributes (e.g., “open,” “on,” “empty”) using CLIP~\citep{clip}-based image-text similarity against a predefined list of possible states. The accompanying depth image is used to project the segmented RGB image into a 3D semantic point cloud, which enables geometric reasoning about object locations. Using spatial heuristics over this point cloud, we extract pairwise inter-object relationships from a fixed set of predicates (e.g., on top of, inside, to the left of, near). The robot’s gripper state is used to determine robot-object interactions (e.g., ``held by robot''). The resulting graph $G_i = (V_i, E_i)$ includes nodes for detected objects, their inferred states, and spatial relationships.

To support subtask-level reasoning, we include a dedicated subtask node representing the current subtask $a_i$. While this node is not connected to other nodes in the graph, it serves as contextual metadata that conditions downstream reasoning about whether the current scene satisfies the expected preconditions for $a_i$. 

\myparagraph{Buffer Construction from Reference Demonstrations}. In order to evaluate whether the environment satisfies the conditions for subtask execution, we infer expected scene graphs from target states derived from successful reference demonstrations. These demonstrations are retrieved based on the target precondition state $P_i$, which may originate from successful trajectories, manually annotated labels, or LLM-generated commonsense descriptions. To capture intra-task variation, where the same task is performed in environments with differing layouts, the system may use successful demonstrations collected from multiple scene configurations. The retrieved subtasks do not necessarily originate from the same overall task that the robot is currently executing, but rather represent valid configurations of the same subtask step across diverse contexts. By referencing similar actions observed in different tasks or environments, the system can leverage a wider range of examples that capture consistent structural and semantic patterns. For instance, a subtask such as ``Pick up mug'' may appear in multiple tasks with different goals, yet exhibit comparable object arrangements. This allows the system to assess whether the current situation aligns with familiar patterns of success, even when the overarching task or settings differ.

% In our setting, we utilize a dataset of successful reference trajectories $\mathcal{D}$, where each trajectory $\tau \in \mathcal{D}$ consists of a sequence of subtasks. These trajectories include successful instances of $a_i$ drawn from a variety of tasks and spatial settings, enabling the system to handle a range of execution contexts while grounding its reasoning in familiar patterns of success. At each step $t_i$, we extract and store the corresponding RGB-D observation $I_i$ as a visually grounded representation of the expected condition for executing $a_i$. These extracted observations are then collected into a buffer of target conditions, which serves as a reference against which the robot compares its current observations during execution.

% In our setting, we leverage a dataset of successful reference trajectories $D$, where each trajectory $\tau \in D$ demonstrates how a high-level task $T$ proceeds under ideal conditions. From these, we extract subtask-specific precondition states, providing visually grounded expectations against which the robot can compare its current observations during execution.

% During runtime, for each subtask $a_i$, we extract expected scene graphs $\{\hat{G}_i^1, \hat{G}_i^2, \dots, \hat{G}_i^k\}$, each corresponding to the starting condition of $a_i$ across demonstrations.

\setlength{\tabcolsep}{30pt} % Adjust column padding
\begin{table*}[t]
    \centering
    \caption{Results of proactive replanning compared to alternative methods. (SR: Success Rate, TET: Total Execution Time in seconds)}
    \resizebox{\textwidth}{!}{
    \begin{tabular}{llcc}
        \toprule
        \textbf{Method} & \textbf{Type} & \textbf{SR (\%)} $\uparrow$& \textbf{TET (sec)} $\downarrow$ \\
        \midrule
        REFLECT~\citep{reflect} & Offline Replanning \textit{after} entire task execution  & 20 & 151.0 \\
        REFLECT-online & Online Replanning \textit{after} failure detected & 30 & 133.9 \\\midrule
        \rowcolor{gray!20} Ours & Online Replanning \textit{before} failure materializes & 70 & 109.2 \\\midrule
        Baseline (w/o Replanning) & & 0 & 84.6 \\
        \bottomrule
    \end{tabular}
    }
    \vspace{-1em}
    \label{tab:proactive_replanning}
\end{table*}

\myparagraph{Context-Aware Reference Retrieval}. Given the full set of successful demonstrations stored in the buffer, the system first filters reference candidates by computing the CLIP-based similarity between the current subtask  those from successful demonstrations. This step ensures that only semantically relevant reference subtasks are selected. The corresponding trajectories from which these subtasks were drawn are then used as reference examples for failure detection. To evaluate the feasibility of executing $a_i$, the robot compares $G_i$ against each of the selected reference graphs ${\hat{G}_i^1, \hat{G}_i^2, \dots, \hat{G}_i^k}$ using a structural similarity metric that measures alignment between the current and expected scene.

\myparagraph{Graph-Based Discrepancy Analysis for Failure Detection.} To assess whether the current environment satisfies the expected conditions for subtask execution, we compare the observed scene graph $G_i^{\text{obs}}$ with the expected graph $G_i^{\text{exp}}$ using a graph based discrepancy analysis algorithm. Each graph consists of a set of object nodes, their associated attributes (e.g., object states), and labeled edges denoting spatial or functional relationships.

% Edge similarity $S_{\text{edge}}$ measures how well spatial or functional relationships align between the graphs. It is defined as the ratio of correctly matched edges to the total number of unique edges:

% \begin{equation}
%     S_{\text{edge}} = \frac{|\mathcal{E}_{\text{matched}}|}{|\mathcal{E}_{\text{exp}} \cup \mathcal{E}_{\text{obs}}|}
% \end{equation}

% Topological similarity $S_{\text{topo}}$ compares the graph structures by measuring the difference in the number of connections for each matched node. A larger degree mismatch lowers the score:

% \begin{equation}
%     S_{\text{topo}} = 1 - \frac{1}{N} \sum_{i=1}^{N} \frac{|\deg(v_i^{\text{obs}}) - \deg(v_i^{\text{exp}})|}{D}
% \end{equation}

Node similarity $S_{\text{node}}$ is computed as the average cosine similarity between matched object nodes, using either CLIP embeddings or semantic segmentation features that encode both object class and state. To penalize extra or missing nodes, the sum is normalized by the total number of unique nodes across both graphs:
\begin{equation}
    S_{\text{node}} = \frac{1}{|\mathcal{V}_{\text{exp}} \cup \mathcal{V}_{\text{obs}}|} \sum_{(v_i^{\text{obs}}, v_i^{\text{exp}})} \cos\left(f(v_i^{\text{obs}}), f(v_i^{\text{exp}})\right)
\end{equation}
Edge similarity ($S_{\text{edge}}$) and structural similarity ($S_{\text{struc}}$) provide complementary views of structural alignment between graphs. 
Specifically, $S_{\text{edge}}$ measures how well spatial or functional relationships align by computing the ratio of correctly matched edges to the total number of unique edges, while $S_{\text{struc}}$ assesses the consistency of node connectivity by measuring differences in node degrees across matched pairs. The two metrics are defined as:
{\footnotesize
\begin{equation}
    S_{\text{edge}} = \frac{|\mathcal{E}_{\text{matched}}|}{|\mathcal{E}_{\text{exp}} \cup \mathcal{E}_{\text{obs}}|} 
    \quad , \quad 
    S_{\text{struc}} = 1 - \frac{1}{N} \sum_{i=1}^{N} \frac{|\deg(v_i^{\text{obs}}) - \deg(v_i^{\text{exp}})|}{D}
\end{equation}
}

We compute a graph similarity score $S$ by averaging the three normalized components:
\begin{equation}
    S = \text{avg}(S_{\text{node}}, S_{\text{edge}}, S_{\text{struc}})
\end{equation}

When the similarity scores \( S_1, S_2, \ldots, S_n \) between the current scene graph \( G_i \) and expected scene graphs \( \{ \hat{G}_i^1, \hat{G}_i^2, \ldots, \hat{G}_i^n \} \) fall below a predefined threshold (i.e., \( S_j < 0.9 \)), the system infers that the current subtask is unlikely to succeed and proactively triggers replanning.

% When the similarity score $S$ between the observed and expected scene graphs falls below a threshold (e.g., $S < 0.9$), the system predicts that the current subtask’s preconditions are likely violated and initiates proactive replanning.

\myparagraph{Replanning Strategy.} Following the detection of a potential failure, our framework decomposes the replanning process into two modular components: a reasoning module and a replanning module, enabling the robot to proactively adjust its plan before executing a subtask that may otherwise result in failure, as illustrated in Figure~\ref{fig:overall}.

The reasoning module is responsible for interpreting the cause of failure. In cases where the comparison to expected scene graphs fail to exceed the similarity threshold, natural language descriptions of the observed and expected scene graphs, along with the task goal and current subtask context, are passed to a language-based reasoning model (i.e., GPT-4o). The model is prompted to explain why the mismatch may lead to task failure and what changes must be made to the current situation or plan to recover from the failure.
    
The output of the reasoning module is then passed to the replanning module, which is subsequently invoked to generate recovery plans. This component takes as input the robot’s current state, the list of available high-level actions, the observable objects in the scene, and the reasoning-informed constraints or suggestions. The replanning model uses this information to generate a revised action sequence that corrects the issue and restores progress toward the task goal. The new plan is then executed online, enabling the robot to dynamically recover from unexpected deviations or environmental constraints.

\section{Experiments}
\label{sec:exp}

\begin{figure*}[h]
    \centering
    \includegraphics[width=1.0\linewidth]{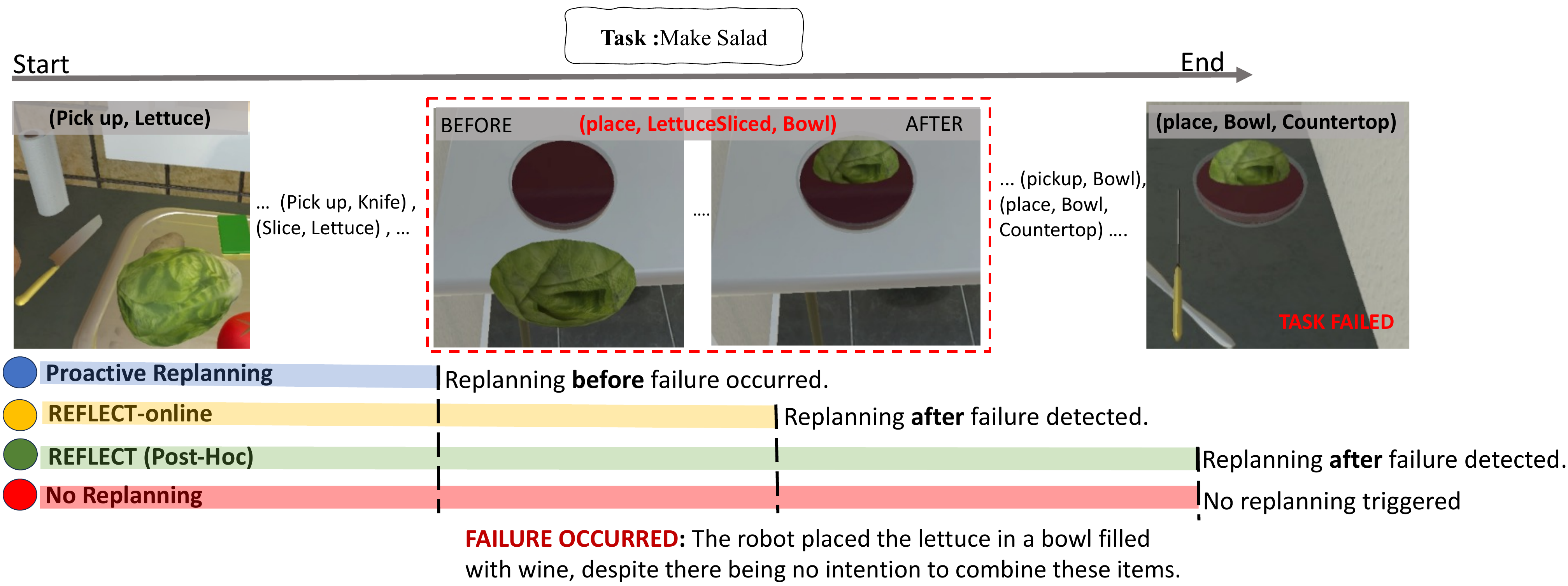}
    % \vspace{-2.0em}
    \caption{ Illustration of task execution timelines across four planning strategies in a failure-prone “Make Salad” task. The robot is expected to place sliced lettuce into a bowl. However, in the failure scenario, the bowl is already filled with wine, an unanticipated condition. Our proactive replanning method anticipates this failure and triggers a correction before the action is executed. In contrast, REFLECT-online detects the failure only after it has occurred, while REFLECT (Post-Hoc) replans only after completing the full task. The no-replanning baseline proceeds without any intervention.}
    \label{fig:failure-detection-reflect}
    \vspace{-1.5em}
\end{figure*}

\myparagraph{Experimental Pipeline.} To evaluate the effectiveness of our proactive replanning framework, we conduct experiments using the RoboFail dataset~\citep{reflect}, which provides a rich set of long-horizon manipulation tasks accompanied by diverse failure scenarios. We select all complex tasks (e.g., make coffee, boil water) composed of semantically meaningful subtasks such as grab mug, place mug on coffee machine, and open drawer, each requiring accurate perception and context-aware decision-making at every stage. More information about each task can be found in the Appendix.

All experiments are conducted in the AI2-THOR simulator~\citep{kolve2017ai2}, which provides high-fidelity environments for household manipulation tasks. For language-based reasoning, we use GPT-4o~\citep{achiam2023gpt} across all methods.

To support failure detection and replanning, we use a set of successful reference trajectories $\mathcal{D}$, where each trajectory $\tau \in \mathcal{D}$ consists of subtasks ${a_1, a_2, \dots, a_n}$ corresponding to high-level actions (e.g., pick up mug). For each task, we collect four successful demonstrations across diverse environments to capture intra-task variation. The reference set also includes cross-task demonstrations from different high-level tasks that may contain  similar instances of the subtask $a_i$. This enables the system to retrieve relevant references from both identical and differing task contexts, allowing the system to handle a range of execution contexts while grounding its reasoning in familiar patterns of success. For each subtask, we store the RGB-D observation $I_i$ captured just before execution, providing visually grounded references for evaluating the current scene.

\myparagraph{Effects of Proactive Replanning.} To evaluate the effectiveness of our proposed proactive replanning method, we construct a dedicated benchmark by extending the RoboFail dataset. These scenarios are specifically designed for this particular experiment to test the robot’s ability to anticipate irreversible failures during long-horizon task execution. We assess performance using task success rate (SR) and Total Execution Time (TET), which measures the average total duration of task execution including time spent on replanning for all tasks. Our method is compared against two baselines that rely on post-hoc replanning with language-based reasoning. The first follows the original REFLECT~\citep{reflect} framework, where replanning is triggered only after the entire task has been completed. The second adapts REFLECT to an online setting, performing verification at the end of each subtask to detect failures. Both baselines are reactive, differing only in whether replanning is triggered at the end of the task or during execution. 

As shown in \Cref{tab:proactive_replanning}, our proactive approach outperforms baseline methods in failure detection and task success. Figure~\ref{fig:failure-detection-reflect} illustrates this contrast in a representative failure case from the “Make Salad” task. Additional results are provided in the Appendix.

\begin{figure*}[h]
    \centering
    \includegraphics[width=1.0\linewidth]{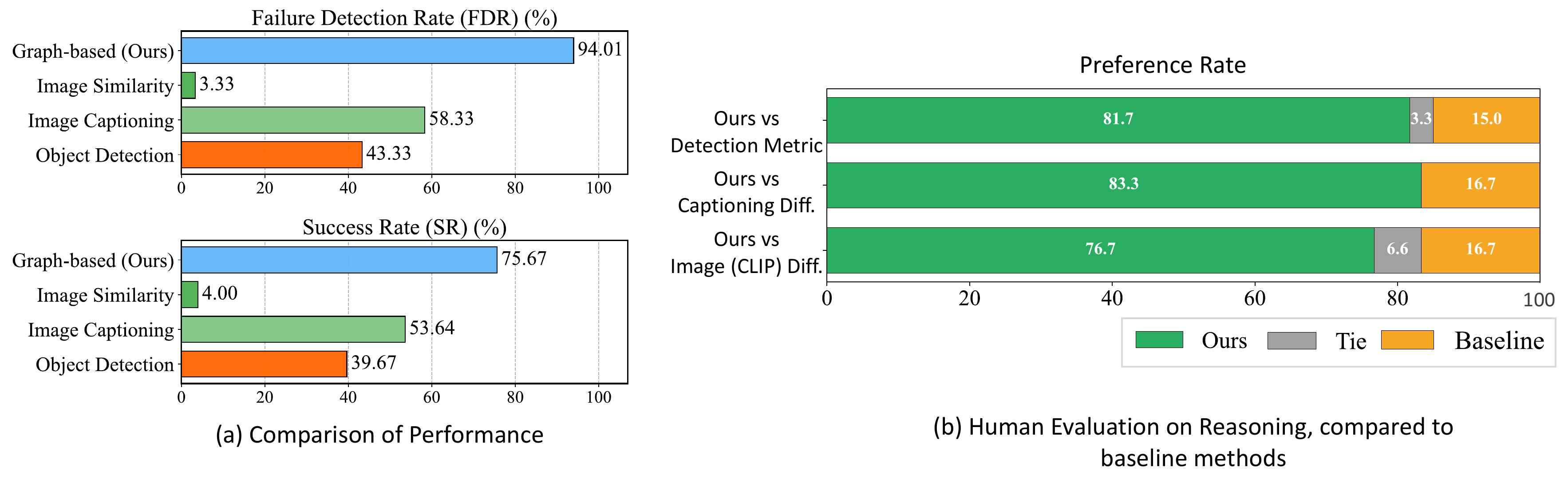}
    \vspace{-2.0em}
    \caption{(a) Comparison of failure detection and task success rates across different approaches. (b) Human evaluation study comparing our scene graph-based failure detection explanations against three baselines: object detection, image captioning, and CLIP image similarity.}
    \vspace{-1em}
    \label{fig:failure-detection}
\end{figure*}

\label{results:faildet}
\begin{figure}[t]
    \centering
    \includegraphics[width=0.9\linewidth]{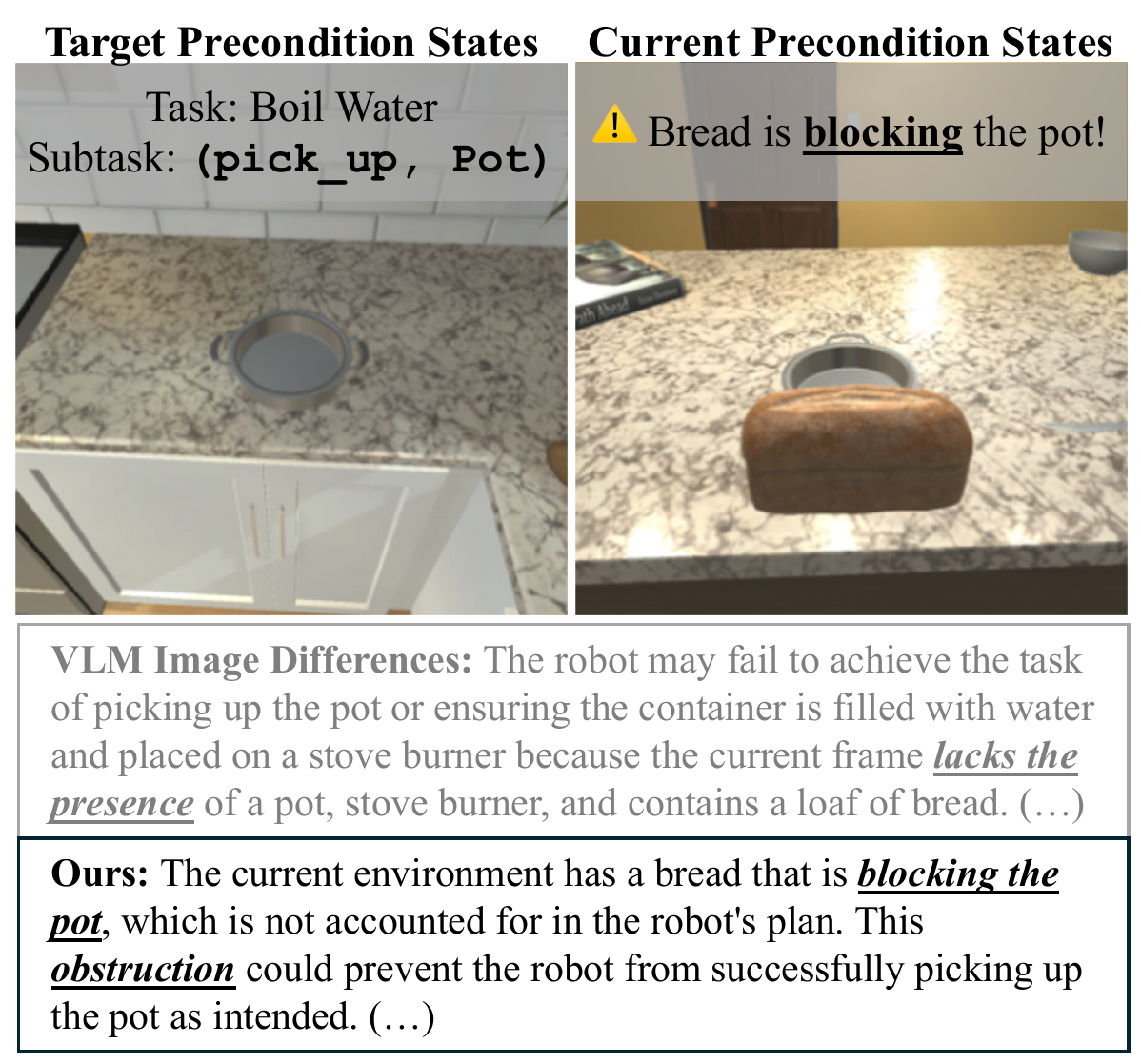}
    \caption{Example failure reasonings from VLM-based method and ours.}
    \vspace{-1em}
    \label{fig:qualitative}
\end{figure}

\myparagraph{Analysis of Potential Failure Detection}. We empirically compare our scene graph-based method against three alternatives: image-level, caption-based, and object-level comparison to validate its effectiveness. Each method receives the same RGB-D observation at the beginning of a subtask and assesses whether the current scene aligns well with the expected conditions derived from reference trajectories. If no expected configuration yields a similarity score above a predefined threshold, the method flags the scene as a potential failure case.

To ensure a fair comparison, we evaluate all methods on the full set of 100 failure scenarios provided by the RoboFail dataset~\citep{reflect}, which offers a rich set of long-horizon  tasks accompanied by diverse failure scenarios. We vary the similarity threshold values (90\%, 85\%, and 80\%) and report the average failure detection rates (FDR) and task success rates (SR). Comprehensive results for all thresholds are in the Appendix. 

 The evaluation results in \Cref{fig:failure-detection}(a) demonstrate that our scene graph-based approach consistently achieves higher failure detection rates (FDR) and task success rates (SR) compared to all baseline methods. In contrast, approaches relying solely on visual or semantic cues, such as CLIP~\citep{clip} embeddings or captioning, demonstrate substantially lower performance, as they emphasize semantic abstraction without capturing relational structure. While the object detection-based method provides object-level recognition, it lacks the spatial reasoning necessary to evaluate the relational configurations critical for subtask feasibility. These results highlight the importance of leveraging structured visual input to enable explicit symbolic and spatial reasoning for reliable proactive replanning.
 
%\label{results:reasoning}
\myparagraph{Analysis of Anticipatory Failure Reasoning.} Building on the same experimental setup, we further evaluate the effectiveness of our framework in reasoning about failure cases.
To ensure a fair comparison, we take episodes where all methods correctly identified a failure and evaluate their ability to explain the cause of that failure. For each method, we generate a natural language explanation using a LLM (GPT-4o)~\citep{achiam2023gpt}. In our approach, the LLM is prompted with structured scene graph differences, whereas for the baselines, it is prompted with either raw image embeddings, generated captions, or detected object-level features.
Figure~\ref{fig:qualitative} further illustrates how spatial context is essential for accurately identifying failure causes. Additional examples are in the Appendix.

To assess explanation quality, we conduct a user study with 15 human evaluators. 
In each case, evaluators are provided with the ground-truth failure cause and two candidate explanations (one generated by our method and one by a baseline method) and corresponding visual observations. 
% Specifically, they are shown (i) the expected scene (what the robot should observe before executing the subtask), (ii) the actual scene (what the robot observes in the perturbed environment), and (iii) the ground-truth reason for failure.
Annotators are instructed to compare the two explanations and select the one better aligned with the ground-truth reasoning. If they find both explanations equally valid or are unsure, they are allowed to select both. More information about this is in Appendix.
As shown in Figure~\ref{fig:failure-detection}(b), our method significantly outperforms all baselines in explanation quality, underscoring the critical role of spatial reasoning in failure understanding. 

% \begin{figure*}
%     \centering
%     \includegraphics[width=0.8\linewidth]{figures/humaneval_bar_v2.pdf}
%     % \vspace{-2.0em}
%     \caption{Human evaluation on reasoning, compared to baseline methods.}
%     \label{fig:human_eval}
% \end{figure*}

% As shown in Figure~\ref{fig:human_eval}, our method significantly outperforms all baselines in explanation quality, underscoring the critical role of spatial reasoning in failure understanding. Figure~\ref{fig:qualitative} further illustrates how spatial context is essential for accurately identifying failure causes. Additional examples are provided in~\Cref{app:additional_reasoning_spatial}.
% \begin{figure*}
%     \centering
%     \includegraphics[width=0.8\linewidth]{figures/humaneval_bar_v3.pdf}
%     % \vspace{-2.0em}
%     \caption{Human evaluation on reasoning, compared to baseline methods.}
%      \vspace{-1.0em}
%     \label{fig:human_eval}
% \end{figure*}

% \begin{figure}
%     \centering
%     \includegraphics[width=1.0\linewidth]{figures/humaneval_bar_v4.pdf}
%     % \vspace{-2.0em}
%     \caption{Human evaluation on reasoning, compared to baseline methods.}
%      \vspace{-1.0em}
%     \label{fig:human_eval}
% \end{figure}

% \input{tables/ablation_full}
%\subsection{Ablation Studies}
%\label{results:ablation}
%In order to understand the contribution of each component in our framework, we conduct an ablation study focusing on failure detection and task success rates. 

\myparagraph{Ablation Studies.} We conduct ablation studies to evaluate the contribution of each component in the scene graph comparison process. As shown in Table~\ref{tab:ablation_faildet} (left), removing the subtask node leads to a moderate drop in failure detection performance, indicating the importance of task context. Excluding node and edge matching causes larger declines, demonstrating that both object identity and relational structure are critical for accurately identifying failures. The full model achieves the highest failure detection rate of 94.01\%.

Additionally, Table~\ref{tab:ablation_faildet} (right) shows that removing the reasoning module results in a substantial drop in task success rat from 75.67\% to 37.67\%. This highlights the module’s essential role in interpreting failures and generating context-aware recovery plans. Without it, the system resorts to uninformed replanning strategies that often fail to address the root causes of failure.

\begin{table}
\centering
\vspace{-1em}
\setlength{\tabcolsep}{0.2cm}
\caption{
Results of ablation studies.
%Ablation results on failure detection (FDR: Failure Detection Rate)
}
{\scalebox{0.9}{
    \begin{tabular}{l|c}
        \toprule
        \textbf{Failure Detection} & \textbf{FDR (\%) $\uparrow$} \\
        \midrule
        \rowcolor{gray!20} Ours & \textbf{94.01} \\
        \midrule
        w/o Subtask Node & 84.33 \red{\scriptsize(8.64\%$\downarrow$)} \\
        w/o Structural Matching & 82.67 \red{\scriptsize(10.3\%$\downarrow$)} \\
        w/o Node Matching & 74.67 \red{\scriptsize(18.3\%$\downarrow$)} \\
        w/o Edge Matching & 70.33 \red{\scriptsize(22.6\%$\downarrow$)} \\
        \bottomrule
        \bottomrule
        \textbf{Task Completion} & \textbf{SR (\%) $\uparrow$} \\
        \midrule
        \rowcolor{gray!20} Ours & \textbf{75.67}\\
        \midrule
        w/o Reasoning & 37.67 \red{\scriptsize(36.0\%$\downarrow$)}\\ \bottomrule
\end{tabular}
}}
\label{tab:ablation_faildet}
\vspace{-1em}
\end{table}

% \input{tables/ablation_sidebyside}

% \vspace{-1em}
\section{Conclusion}
\label{sec:conclusion}
We propose a novel proactive replanning framework that detects and mitigates potential failures before they occur during robotic task execution by grounding decisions in structured visual understanding. By leveraging state discrepancy analysis through scene graph comparisons between the current observation and reference trajectories extracted from successful demonstrations, our method enables pre-execution detection with online replanning without relying on dense annotations or frequent LLM calls. Experimental results on RoboFail benchmark demonstrate that our approach significantly outperforms prior baselines in both failure detection and task success rates, highlighting the critical importance of explicit symbolic and spatial reasoning for robust autonomy. Overall, we suggest that proactive, structured replanning offers a promising path forward for enabling safe, adaptive, and efficient robot behavior in complex environments.
{
    \small
    \bibliographystyle{ieeenat_fullname}
    \bibliography{main}
}

\end{document}